%% file: main.tex
\documentclass{article}

\usepackage[final]{neurips_2019}

\usepackage[utf8]{inputenc} 
\usepackage[T1]{fontenc}    
\usepackage{hyperref}       
\usepackage{url}            
\usepackage{booktabs}       
\usepackage{amsfonts}       
\usepackage{nicefrac}       
\usepackage{microtype}      

\usepackage{subcaption}
\usepackage{graphicx}
\usepackage{xspace}

\usepackage{natbib}

\def\Fig#1{Fig.~\ref{fig:#1}}

\usepackage{listings}
\usepackage{color}
\usepackage[dvipsnames]{xcolor}

\definecolor{mygreen}{rgb}{0,0.6,0}
\definecolor{mygray}{rgb}{0.5,0.5,0.5}
\definecolor{mymauve}{rgb}{0.58,0,0.82}
\definecolor{deepblue}{rgb}{0,0,0.5}
\definecolor{deepred}{rgb}{0.6,0,0}
\definecolor{deepgreen}{rgb}{0,0.5,0}

\newcommand\YAMLcolonstyle{\color{red}\mdseries}
\newcommand\YAMLkeystyle{\color{black}\bfseries}
\newcommand\YAMLvaluestyle{\color{blue}\mdseries}

\makeatletter

\newcommand\language@yaml{yaml}

\expandafter\expandafter\expandafter\lstdefinelanguage
\expandafter{\language@yaml}
{
	keywords={true,false,null,y,n},
	keywordstyle=\color{darkgray}\bfseries,
	basicstyle=\YAMLkeystyle,                                 
	sensitive=false,
	comment=[l]{\#},
	morecomment=[s]{/*}{*/},
	commentstyle=\color{purple}\ttfamily,
	stringstyle=\YAMLvaluestyle\ttfamily,
	moredelim=[l][\color{orange}]{\&},
	moredelim=[l][\color{magenta}]{*},
	moredelim=**[il][\YAMLcolonstyle{:}\YAMLvaluestyle]{:},   
	morestring=[b]',
	morestring=[b]",
	literate =    {---}{{\ProcessThreeDashes}}3
	{>}{{\textcolor{red}\textgreater}}1     
	{|}{{\textcolor{red}\textbar}}1 
	{\ -\ }{{\mdseries\ -\ }}3,
}

\newcommand\yamlstyle{\lstset{
		language=yaml,
		basicstyle=\scriptsize\ttfamily,        
		breakatwhitespace=false,         
		breaklines=true,                 
		keywordstyle=\color{blue},       
		otherkeywords={self},             
		keywordstyle=\color{deepblue},
		emph={pipeline, priority,type, streams,test,task},          
		emphstyle=\color{deepred},    
		stringstyle=\color{deepgreen},
		commentstyle=\color{mygreen},    
		showstringspaces=false,            %
}}
\lstnewenvironment{yaml}[1][]
{
\yamlstyle
\lstset{#1}
}
{}

\title{PyTorchPipe: a framework for rapid prototyping of pipelines combining language and vision}

\author{Tomasz Kornuta \\
  IBM Research AI\\
  Almaden Research Center\\
 San Jose, CA 15120 \\
  \texttt{tkornut@us.ibm.com}
}

\begin{document}

\maketitle

\begin{abstract}
Access to vast amounts of data along with affordable computational power stimulated the reincarnation of neural networks.
The progress could not be achieved without adequate software tools, lowering the entry bar for the next generations of researchers and developers.
The paper introduces PyTorchPipe (PTP), a framework built on top of PyTorch.
Answering the recent needs and trends in machine learning, PTP facilitates building and training of complex, multi-modal models combining language and vision (but is not limited to those two modalities).
At its core, PTP employs a component-oriented approach and relies on the concept of a pipeline, defined as a directed acyclic graph of loosely coupled components.
A user defines a pipeline using yaml-based (thus human-readable) configuration files, whereas PTP provides generic workers for their loading, training, and testing using all the computational power (CPUs and GPUs) that is available to the user.
The paper covers the main concepts of PyTorchPipe, discusses its key features and briefly presents the currently implemented tasks, models and components.

\end{abstract}

\input{intro}

\input{pytorchpipe_description}

\input{application}


\bibliographystyle{abbrvnat}

\input{main.bbl}
\end{document}

%% file: intro.tex
\section{Introduction}

Deep learning~\citep{lecun2015deep} enabled impressive improvement in performance across many problem domains, and neural network based models became dominating solutions, achieving state of the art performances, e.g., in speech recognition~\citep{graves2013speech}, 
image classification~\citep{krizhevsky2012imagenet},  object detection~\citep{redmon2016you}, instance segmentation~\citep{he2017mask},
question answering~\citep{weston2014memory}
or machine translation~\citep{bahdanau2014neural}.
This would not be possible without (1) access to large-scale datasets, (2) harnessing the computational power and parallel processing of GPUs and (3) the appropriate software.
Starting from low-level libraries such as CUDA~\citep{sanders2010cuda}, the AI community has been developing tools and libraries offering more and more levels of abstractions above the multiply–accumulate operations and elementary tensor algebra, e.g.\ Torch~\citep{collobert2002torch}, Theano~\citep{bastien2012theano}, Chainer~\citep{tokui2015chainer} or TensorFlow~\citep{abadi2016tensorflow}.
Those API based tools facilitate building and training of deeper and more complex neural models and significantly lower the entry bar for new users and researchers.
As the community started to shift towards multi-problem suites such as 
MS COCO~\citep{lin2014microsoft}, being a large-scale object detection, segmentation, and captioning dataset, or ParlAI~\citep{miller2017parlai} and AllenNLP~\citep{Gardner2017AllenNLP}, offering wide varieties of linguistic tasks,
researchers started to feel the need for more specialized APIs that will enable comparison of different models under exactly the same settings and target the so-called \textit{reproducibility crisis}~\citep{hutson2018artificial}, 
with examples being object detection APIs such as Detectron~\citep{Detectron2018} for PyTorch~\citep{paszke2017automatic} or object detection API in TensorFlow~\citep{huang2017speed}. 
The need for more abstraction from both problem domain (dataset) and model specialized for a given domain led to development of the next generation of tools, 
such as Tensor2Tensor~\citep{vaswani2018tensor2tensor}, MI-Prometheus~\citep{kornuta2018accelerating}, Pytia~\citep{singh2018pythia}, OpenSeq2Seq~\citep{kuchaiev2018mixed} or Ludwig~\citep{Molino2019}.
Aside of the mentioned modularity and abstraction, all those tools offer powerful scripts, agnostic to dataset and model, that standardize training (and testing) procedures, once again lowering the entry bar and enabling the user to focus more on the model rather than on the training loop, validation procedures, utilization of GPUs etc.

\begin{figure}[htbp]
    \centering
    \includegraphics[width=\textwidth]{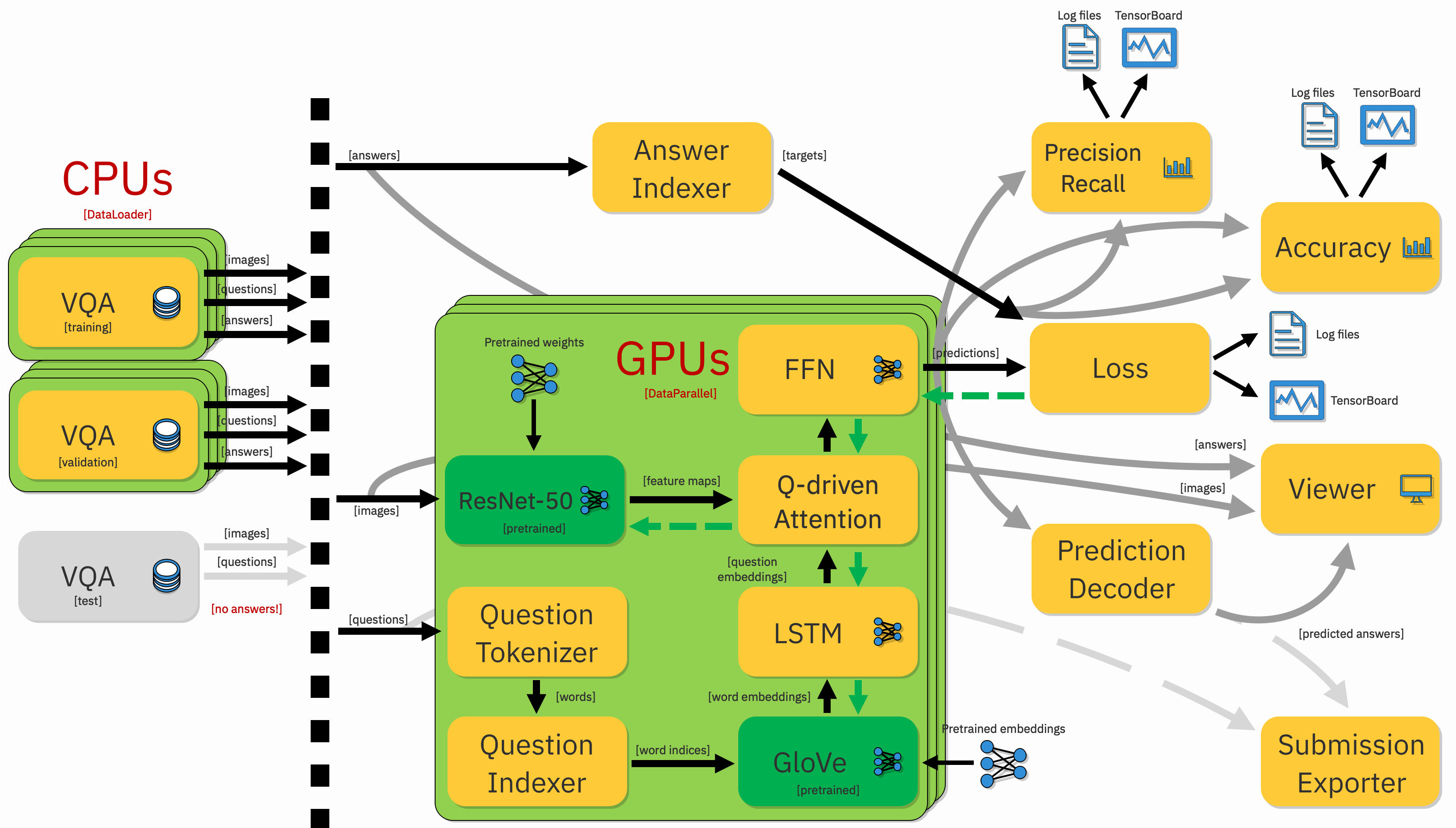}
    \caption{Exemplary multi-modal flow diagram. In PTP each block is implemented as a separate component, whereas framework handles passing data between them and controls the execution order.}
    \label{fig:exemplary_multi_modal_diagram}
\end{figure}

As the individual problem domains matured and saturated with diverse solutions, a new strong trend 
called Multi-Modal Machine Learning~\citep{baltruvsaitis2018multimodal} emerged, with tasks combining vision and language~\citep{mogadala2019trends} such as Image Captioning~\citep{karpathy2015deep} and Image/Visual Question Answering (VQA)~\citep{malinowski2014multi,antol2015vqa} becoming slightly more popular than the others.
A clearly different nature of those two modalities (spatial vs sequential) implies diverse transformation that the inputs need to undergo, e.g. convolutional layers in the case of images and tokenization, indexing, word embeddings etc. in the case of questions.
As a result, the most general VQA architecture consists of four major modules:  two encoders responsible for encoding of raw image and question into more useful representations, a reasoning module that combines them and decoder that produces the answer.
In early VQA systems, reasoning modules implemented diverse multi-modal fusion mechanisms~\citep{malinowski2018visual}, from concatenation to pooling, e.g. MCB~\citep{fukui2016multimodal} and MLB~\citep{kim2017hadamard}, to diverse (co-)attention mechanisms, e.g. question-driven attention over image features~\citep{kazemi2017show}.
More recently, researchers have focused on reasoning mechanisms such as Relational Networks~\citep{santoro2017simple,desta2018object} and Memory, Attention and Composition (MAC) networks~\citep{hudson2018compositional,marois2018transfer}.
Still, they all can fit into the \textit{architectural pattern} defined above.
Similarly, one can assume adequate architectural patterns for Image Captioning or any other task, e.g. Visual Dialog~\citep{das2017visual} or Visual Image Generation~\citep{mogadala2019trends}.
Such assumptions enable development of software facilitating their creation, for example, OpenSeq2Seq assumes model to follow the \textit{Encoder(s)-Decoder} architecture, Ludwig imposes it to be composed of \textit{Encoder(s)-(Combiner)-Decoder}, whereas Pytia assumes that user will create a \textit{monolithic} model, but instead standardizes interfaces by enforcing that batches fetched by all datasets (Tasks) would have exactly the same signature consisting of \textit{(image, context, text, output)}.

However, in practice, some additional computations that need to be done.
When looking at an exemplary VQA system presented in \Fig{exemplary_multi_modal_diagram}, one might notice that answers also undergo some transformations (from words to indices/one-hot encoding) and aside of loss there are some additional modules calculating statistics.
Besides, for the \textit{test split} the answers are typically not provided.
Hence some of the modules should not be executed at all when working with that split.
Finally, it is a common practice~\citep{jiang2018pythia,teney2018tips} to use transfer learning~\citep{pan2009survey} and incorporate into the model image encoders and/or word embeddings  pre-trained a priori on external datasets.
Moreover, typical approach during training is to \textit{freeze} the encoder(s) weights at first (which enables faster convergence), but at the end to \textit{unfreeze} them and fine-tune the whole model jointly (which gives a slight accuracy boost).
Those observations lead to relaxation of assumptions regarding utilization of fixed architectural patterns  and formulated the requirement for flexible decomposition of models into smaller modules that laid the core foundation for PyTorchPipe.





%% file: pytorchpipe_description.tex
\section{PyTorchPipe}

PyTorchPipe (PTP)\footnote{\url{https://github.com/ibm/pytorchpipe}}, being the main contribution of this paper, is a software framework build on top of PyTorch~\citep{paszke2017automatic}.
PTP is a component-oriented framework that facilitates development and rapid prototyping of computational multi-modal pipelines and comparison of diverse neural network-based models.
The most important feature of PTP is the \textbf{decomposition of complex "monolithic" models into graphs with many inter-connected computational nodes}, that increases their reusability while still enabling their joint training.
As a consequence, PyTorchPipe enables:
\begin{itemize}
\item Plugging in/out “modules” (components) \textit{at run-time}\footnote{Run-time is defined here as the moment of execution of the experiment.},
\item Importing the pre-trained models (or their “parts”) and saving them during/after training,
\item Freezing/unfreezing of the models (or their “parts”) on demand (at run-time),
\item Run-time parametrization of all “modules”,
\item Pipeline-agnostic, generic scripts for training and testing, enabling run-time parametrization (hyper-parameters) and utilization of many CPUs/GPUs on-demand.
\end{itemize}
Additionally, PTP offers several other functionalities useful for rapid prototyping and training of new models, such as: different optimizers, sampling methods, automated logging and  statistics collection facilities, the export of statistics to files (and TensorBoard) and automated saving of the best performing pipelines/models during training.
Moreover, PTP provides out-of-a-box library of diverse, parameterizable components that can be used for rapid creation of various prototypes.

\begin{figure}[htbp]
    \centering
    \includegraphics[width=0.9\textwidth]{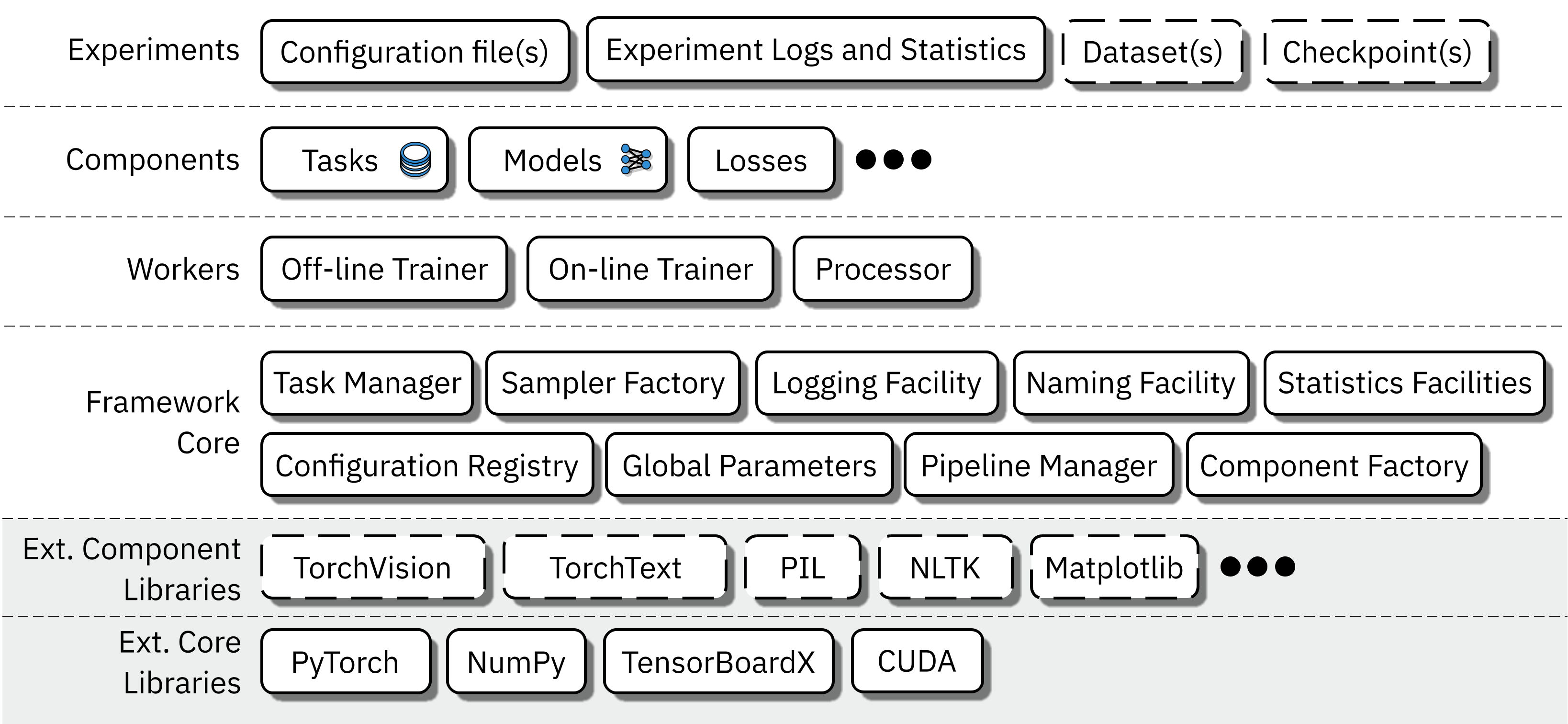}
    \caption{Architecture of the PyTorchPipe framework.}
    \label{fig:layers}
\end{figure}

From the architectural point of view, PTP can be seen as a stack of layers, as presented in \Fig{layers}.
The two bottom layers are external libraries that PTP and its components depend on.
The \textbf{Framework Core} layer contains all the tools and utilities realizing core functionalities, i.e. loading configuration files into registry, managing registry and global parameters, building the pipeline based on loaded configuration, component factories, loggers, statistics collectors and aggregators.
\textbf{Workers} later use those for building and deploying the actual pipelines and execution of experiments.
The \textbf{Components} layer contains the implementations of components along with their default configurations that are loaded and managed by the workers.
Finally, the \textbf{Experiments} layer contains files associated with a given experiment that the user wants to perform, staring from the configuration file(s), all the logs and statistic collected and saved to files during experiment execution, files associated with a given dataset, loaded and saved checkpoint(s) with weights of the models incorporated in the pipeline etc.
Please note that the layer architecture reflects three types of user roles classified by their expertise, from top: 
\begin{itemize}
\item \textbf{Experiment runners} might have little to no knowledge of the underlying models, training procedures, methods (backpropagation) etc. but are able to define experiment by writing configuration file(s) incorporating the existing components and relying on existing workers\footnote{Also note that all PTP components and workers come with default configuration files, so one does not have to look at the actual implementation when defining its own experiment.}.
\item \textbf{Component developers} must understand how a given model works, however, do not need to understand every aspect of the training/testing procedures nor how PTP operates internally.
\item \textbf{Worker developers} must understand both how models and training/testing work, as well as be familiar with  internal operation of PTP core, how pipelines are assembled etc.
\end{itemize}

\subsection{Pipelines}
\begin{figure}[htbp]
  \null\hfill
  \begin{subfigure}{.6\textwidth}
    \includegraphics[width=\textwidth]{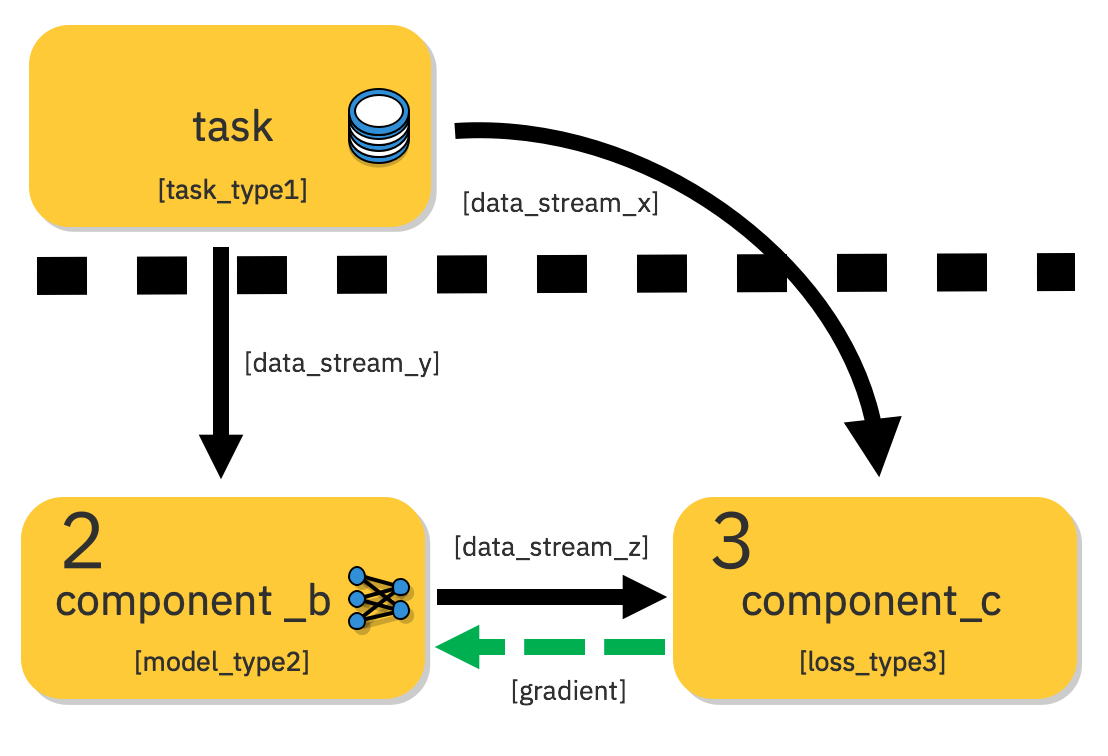}
  \end{subfigure}%
  \hfill
  \begin{subfigure}{.32\textwidth}
    \begin{yaml}
training:
  task:
    type: task_type1
    streams:
      output1: data_stream_x
      output2: data_stream_y
pipeline:
  component_b:
    priority: 2
    type: model_type2
    streams:
      input: data_stream_y
      output: data_stream_z
  component_c:
    priority: 3
    type: loss_type3
    streams:
      input1: data_stream_x
      input2: data_stream_z
    \end{yaml}
  \end{subfigure}%
  \hfill\null
  \caption{Exemplary pipeline (left) with its YAML definition (right).}
  \label{fig:yaml_example}
\end{figure}

What the previously mentioned tools typically define as a \textit{Model}, in PTP is framed as a \textbf{Pipeline}: a directed acyclic graph (DAG) consisting of many inter-connected components.
The components are \textit{loosely coupled} and care only about the \textbf{Input Streams} they retrieve from and \textbf{Output Streams} they publish to.
There are three special classes of components: \textbf{Task}, \textbf{Model} and \textbf{Loss}.
\textbf{Tasks} are components feeding batches of samples into pipelines, and due to the training methodologies that interlace training and validation batches there are treated is a slightly different way, thus are in fact \textit{outside of pipeline}.
\textbf{Models} are components containing trainable weights, whereas \textbf{Losses} are components calculating values of loss function for a given batch (and from which gradients are back-propagated).
Once the components are implemented, a user defines a pipeline by writing a configuration file (\Fig{yaml_example}).
Components are executed following the priorities defined by the user.
A pipeline can consist of any number of components of a given type, including many Models and many Losses (what enables e.g. Multi-Task Learning~\citep{caruana1997multitask}) and other components providing required transformations and computations.
The user connects the components by explicitly indicating the data streams connections; Naming Facilities play an important role here, enabling one to rename any stream at the run time, without the need to change the component code.
The framework offers full flexibility and it is up to the programmer to choose the granularity of his/her components/models/pipelines.

\subsection{Component}
\begin{figure}[htbp]
    \centering
    \includegraphics[width=0.5\textwidth]{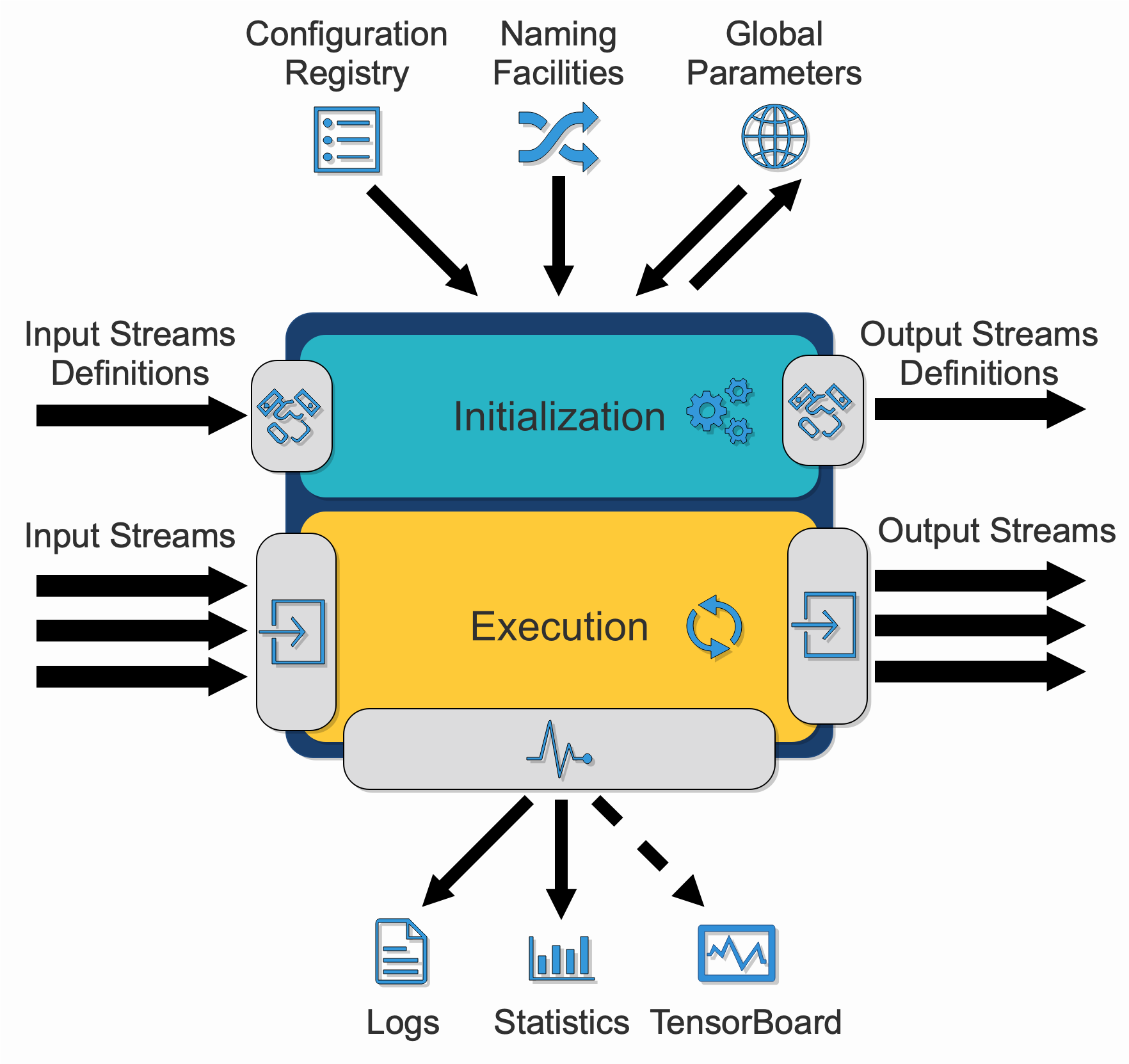}
    \caption{Component operation modes}
    \label{fig:ptp_component}
\end{figure}

All components have two basic operation modes: \textbf{Initialization} and \textbf{Execution} (\Fig{ptp_component}), that also reflect two main modes of PTP workers.
During \textbf{Initialization} workers load configuration file(s) and create all instances of the components in the pipeline.
Each component receives its subsection of configuration file and uses it to set its internal variables, rename stream/global names, get/set global variables, and perform other component-specific operations (load dataset/create nn module etc.). 
As soon as all components are initialized the worker performs \textit{handshaking}, i.e. makes sure that all streams are correctly connected by checking compatibility of their output-input definitions.
Initialization happens only once and when finished, workers switch to \textbf{Execution} mode.
During \textbf{Execution} each component processes data from input streams and publishes results to output streams, logs its operation, collects statistics (automatically exported to CSV files and, optionally, to Tensorboard) etc.

\subsection{Workers}
PTP workers are python scripts that are agnostic to tasks/models/components/pipelines that they are supposed to work with. Currently our framework offers three types of workers:
\begin{itemize}
\item \textbf{ptp-offline-trainer}: a trainer relying on classical methodology interlacing training and validation at the end of every epoch; creates separate instances of training and validation tasks and trains the models by feeding the created pipeline with batches of data.
\item \textbf{ptp-online-trainer}: a flexible trainer creating separate instances of training and validation tasks; validation is performed using a single batch every user-defined interval; relying on the notion of an episode rather than epoch.
\item \textbf{ptp-processor}: performing a single pass over all the samples (batches) returned by a given task; useful for collecting scores on test sets, answers for submissions for competitions etc.
\end{itemize}
In its core, to accelerate the computations on their own, PTP relies on PyTorch and all workers extensively use its mechanisms for distribution of computations on CPUs/GPUs, including multi-process data loaders and multi-GPU data parallelism. The Tasks, Models and other components are agnostic to those operations and the user indicates whether to use the former (data loaders) in configuration files or the latter (GPUs) by passing an adequate argument (\texttt{--gpu}) at run-time.

\subsection{Task/Model/Component Zoo}
\begin{figure}[htbp]
    \centering
    \includegraphics[width=\textwidth]{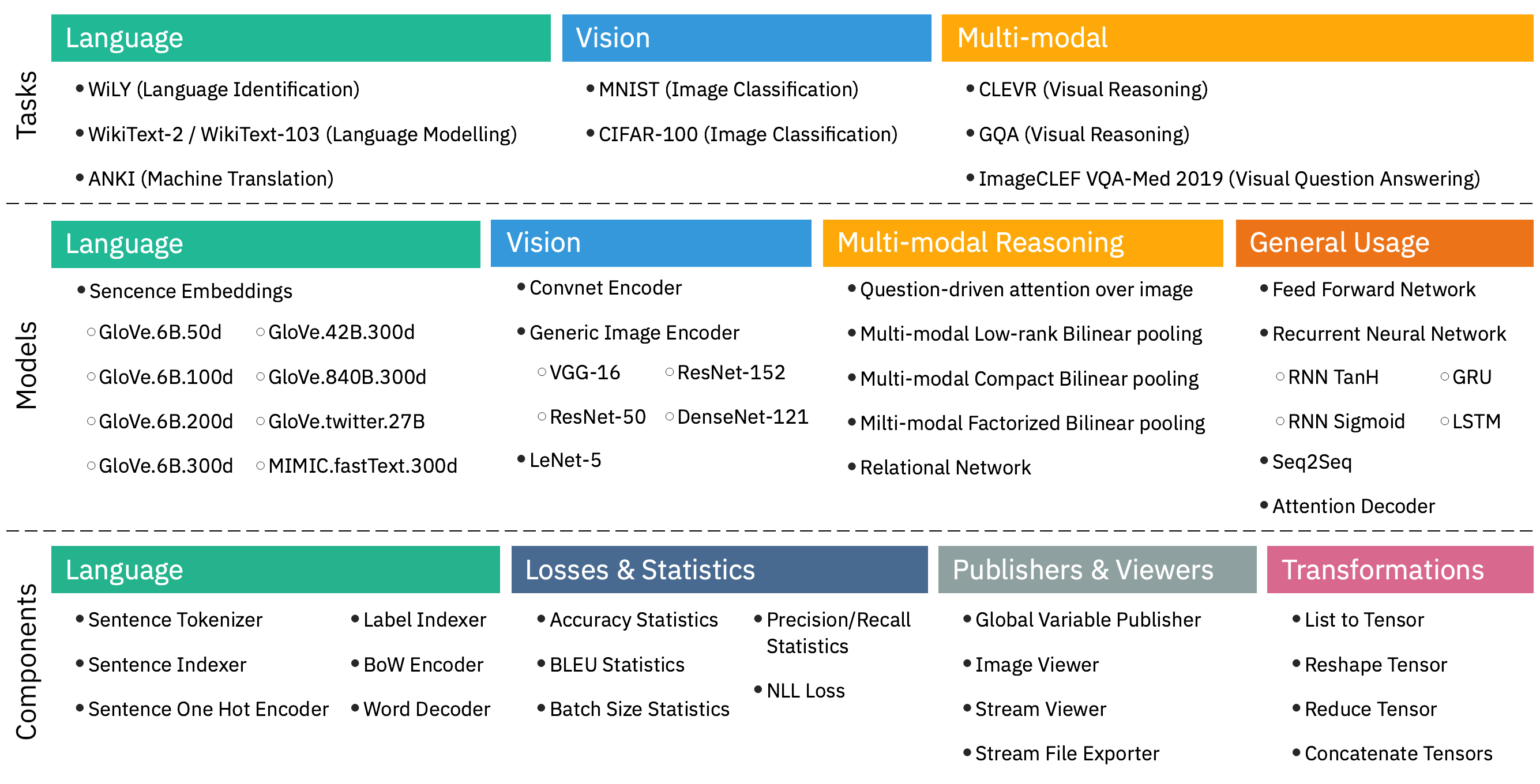}
    \caption{Tasks, models and other components currently available in PTP}
    \label{fig:ptp_tasks}
\end{figure}

PTP is generally agnostic to operations performed by the components and data passed in data streams.
However, the currently provided tasks, models and components focus mostly on applications that can be roughly classified as belonging to: (a) computer vision domain, (b) Natural Language Processing domain and (c) domain combining vision and language.
\Fig{ptp_tasks} reflect this.
Aside of selection of tasks covering various aspects from those three domains, PTP offers several models that are specialized for language or vision, but also includes several general usage components, such as Feed Forward Network (with variable number of Fully Connected layers with activation functions and dropout between them) or Recurrent Neural Network (with several cell types available; with activation functions and dropout; the component can also work as encoder or decoder).
PTP also provides several ready to use non-trainable (but still parametrizable) components.
That list includes components useful when working with language, vision or other types of streams (e.g., tensor transformations).
There are also several general-purpose components, such as components calculating losses and statistics, and publishers and viewers that enable the user to digest and analyze contents of data streams.
The diversity of those components illustrates the flexibility of the framework.

%% file: application.tex
\section{Use case: the Image-CLEF Med-VQA 2019 challenge}

The VQA-Med 2019~\cite{ImageCLEFVQA-Med2019} is an open challenge organized as part of the ImageCLEF 2019 initiative~\cite{ImageCLEF19}.
The associated dataset belongs to the Visual Question Answering (VQA)  problem domain. with a focus on radiology images.
Despite its small size (a training set of 3,200 images with 12,792 question-answer pairs, divided into four C1--C4 categories) the dataset is scattered, noisy and heavily biased, thus 
we decided to use it as a test bed for PTP, as we could encounter all the challenges that one must deal with in practical scenarios.

\begin{figure}[htbp]
    \centering
    \includegraphics[width=\textwidth]{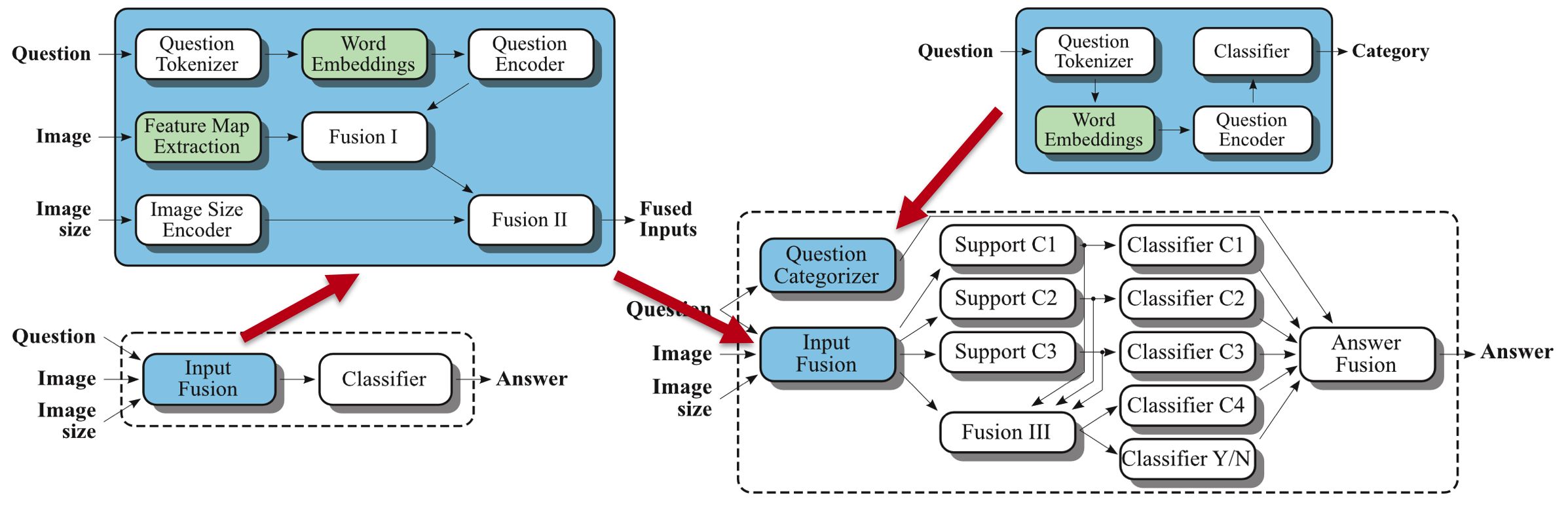}
    \caption{Multi-stage training of the Supporting Facts Network}
    \label{fig:sfn_module_transfer_learning}
\end{figure}


Summarizing our efforts, within a month we have build more than 50 diverse pipelines, developed dozens of components and made 4 entries to the leaderboard by submitting the files generated straight in PTP.
Our final model, called \textbf{Supporting Facts Network} (SFN)~\citep{kornuta2019leveraging}, shared knowledge between upstream and downstream tasks through the use of a coarse-grained categorizer combined with five task-specific, fine-grained classifiers.
Training of the final SFN architecture (the whole final pipeline had 53 components!) was a complex procedure, with several transfer learning steps (represented as red arrows in \Fig{sfn_module_transfer_learning}): (0) using pre-trained "Word Embeddings" (GloVe.6B.50d) and "Feature Map Extractor" (ResNet-50), (1) pre-training of ”Question Categorizer” on C1,C2,C3 and C4 categories, (2) pre-training “Input Fusion” on C1,C2 and C3 categories, (3) loading and freezing the ”Question Categorizer”, loading and fine-tuning the "Input Fusion" and (4) training jointly the final SFN architecture using 5 losses (one for each of the fine-grained classifier).
Despite we haven't won the challenge (ranked 7), we found PTP extremely useful, supporting all the abovementioned operations and enabling us to rapidly prototype new solutions and ideas.


